\documentclass{article}
\usepackage{spconf,amsmath,graphicx}

\usepackage{CJK}
\usepackage{multirow}
\usepackage{makecell}

\usepackage{enumitem}

\title{An Empirical Investigation of Domain Adaptation Ability for Chinese Spelling Check Models}
\name{Xi Wang, \ Ruoqing Zhao, \ Hongliang Dai, \ Piji Li \sthanks{Corresponding author}}
\address{Nanjing University of Aeronautics and Astronautics}
\begin{document}
%
\maketitle
\begin{abstract}
Chinese Spelling Check (CSC) is a meaningful task in the area of Natural Language Processing (NLP) which aims at detecting spelling errors in Chinese texts and then correcting these errors. However, CSC models are based on pretrained language models, which are trained on a general corpus. Consequently, their performance may drop when confronted with downstream tasks involving domain-specific terms. In this paper, we conduct a thorough evaluation about the domain adaption ability of various typical CSC models by building three new datasets encompassing rich domain-specific terms from the financial, medical, and legal domains. Then we conduct empirical investigations in the corresponding domain-specific test datasets to ascertain the cross-domain adaptation ability of several typical CSC models. We also test the performance of the popular large language model ChatGPT. As shown in our experiments, the performances of the CSC models drop significantly in the new domains.
\end{abstract}
\section{Introduction}
\label{sec:intro}

\begin{figure}[!t]
\centering
\includegraphics[width=0.8\columnwidth,height=0.8\columnwidth]{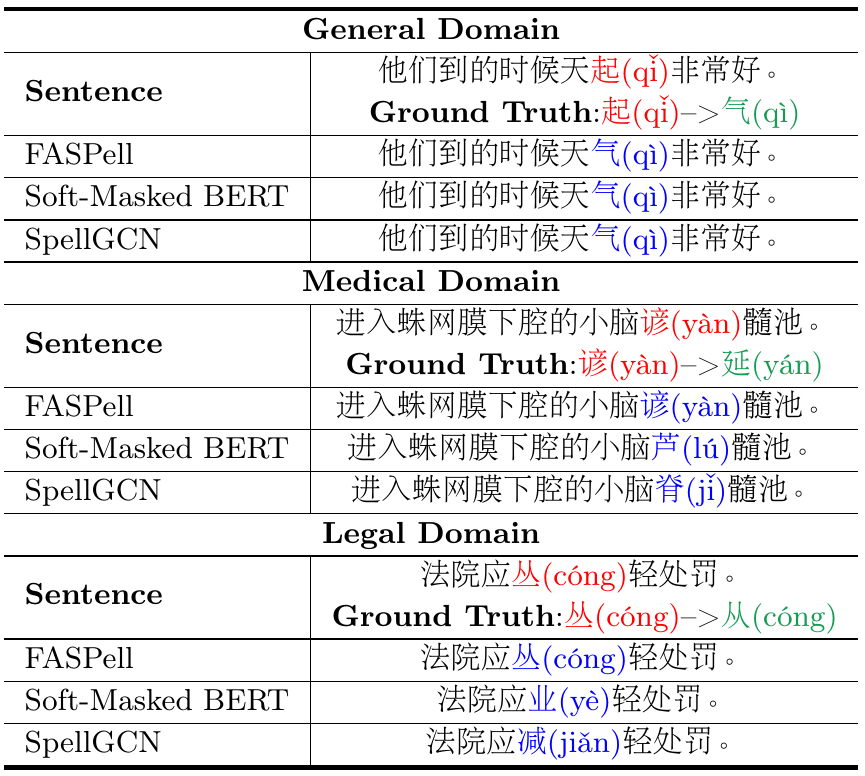}
	\caption{The output of three models in different domains. Wrong characters are marked in red, ground truth characters are highlighted in green, and predicted characters are highlighted in blue.}
	\label{fig}
\vspace{-6mm}
\end{figure}

Chinese Spelling Check (CSC) aims to detect and correct spelling mistakes in Chinese sentences. It is a challenging task in Chinese Natural Language Processing (NLP) and holds substantial importance for various downstream applications, including Optical Character Recognition (OCR)~\cite{afli-etal-2016-using}, Automatic Speech Recognition (ASR)~\cite{gao-etal-2010-large}, AI Writing Assistant~\cite{DBLP:journals/corr/abs-2208-01815}, and numerous other tasks that contain Chinese texts. By employing CSC in downstream tasks, a considerable reduction in common errors caused by both machines and humans can be achieved.

As spelling mistakes in Chinese texts need to be detected in a specific sentence, effective detection and correction of such errors rely on context comprehension of CSC models. Fortunately, great success has been achieved in CSC since pretrained language models became the backbone model. However, the training data of BERT mainly comes from BooksCorpus and Wikipedia\cite{devlin-etal-2019-bert}, while the corpus of downstream tasks vary differently. This mismatch will lead to a considerable decline in performance when testing the CSC model on domain-specific terms. This phenomenon is exemplified in Figure~\ref{fig}, where CSC is conducted using CSC models FASPell~\cite{hong-etal-2019-FASPell}, Soft-Masked BERT~\cite{Softmasked}, and SpellGCN~\cite{spellgcn} across various domains. The results demonstrate that all those three models can obtain correct detection and correction results on the general domain, while performance collapses on medical and legal domains. The models often replacing them with commonly used terms rather than specific terms.

Considering the practical scenarios where input texts may originate from diverse domains, it is crucial and necessary to evaluate the domain adaptation ability of current typical CSC models. As there is no public domain-specific CSC datasets available, it becomes imperative to develop novel datasets to quantitatively assess the domain adaptation ability of classical models. Specifically, we focus on three distinct domains: financial, medical, and legal domain. After corpora collection, we apply predefined rules to introduce errors into normal sentences. Furthermore, we deliberately construct sentences with two error positions, thereby enhancing the challenges of detecting and correcting all the errors.

Furthermore, it is important to note that several existing models~\cite{hong-etal-2019-FASPell,Softmasked,spellgcn,Lv_2022} rely on supervised learning approaches which are trained on datasets with limited coverage of terminologies from specialized domains. As a result, these models may have poor performance when confronted with out-of-domain tasks. This raises concerns about the ability of task-specific models to maintain the domain adaptation capability inherent in the original pretrained language models. To verify this conjecture, we also conduct cross-domain evaluation on some unsupervised CSC methods such as uChecker~\cite{li-2022-uchecker} which can preserve the knowledge of the original pretrained language models as much as possible. With the construction of the aforementioned three domain-specific datasets, we systematically evaluate the typical CSC models mentioned above on these datasets and present the corresponding results.

There are works such as ECSPell~\cite{Lv_2022} that raises the issue of the domain adaptation problem in the CSC area, and evaluates some state-of-the-art models on the human-annotated domain datasets. However, constructing a dataset through human annotation is a time-consuming process, and the proposed approach incorporates external domain-specific knowledge to filter output. While this work serves as an inspiration for our research, it does not consider unsupervised methods and lacks a comprehensive analysis of the compared models.

We are also interested in the performance of large language models in CSC. To this end, we choose ChatGPT as the example and use prompts to make evaluation of the performance of large language models.

In summary, our contributions are as follows: 
\begin{itemize}[topsep=0pt]\setlength\itemsep{-0.5em}
    \item We are the first to propose to investigate the domain adaptation ability of both supervised and unsupervised CSC models and construct an automatically-generated domain-specific dataset for financial, medical, and legal domains respectively to perform the evaluation.
    \item We evaluate various typical supervised and unsupervised CSC models on the generated domain-specific datasets, and find out that the supervised models perform worse than the unsupervised models.
    \item We also test the performance of the popular language model ChatGPT and find its deficiency in CSC.
\end{itemize}

\section{Methods}

\subsection{Task Formulation}
Given a Chinese sentence input $X=\{x_1,x_2,\dots,x_n\}$ whose length is $n$, the CSC models aim to detect and correct spelling errors in $X$, and then output the corrected sentence $Y=\{y_1,y_2,\dots,y_n\}$. Note that $X$ and $Y$ have the same length. The models operate by replacing incorrect words with their corresponding corrected forms, while preserving the remaining words in the original input sentence.

\subsection{Domain-Specific Dataset Constructing}
\label{sec:data_construct}

\begin{table}[!t]
\centering
\resizebox{0.9\columnwidth}{!}{
\begin{tabular}{lcccccc}
\Xhline{3\arrayrulewidth}
\textbf{Dataset} & \textbf{Err\_sen} & \textbf{Total\_sen} & \textbf{Avg\_len} & \textbf{Max\_err} \\ \hline
$\mathrm{Financial}_{test}$ & 734  & 1082  & 70.1 & 8 \\
$\mathrm{Medical}_{test}$   & 2624 & 4100  & 52.7 & 9 \\
$\mathrm{Legal}_{test}$       & 2195 & 3319  & 75.2 & 9 \\
$\mathrm{SIGHAN13}_{test}$  & 971  & 1000  & 74.3 & 5 \\
$\mathrm{SIGHAN14}_{test}$  & 520  & 1062  & 50.0 & 7 \\
$\mathrm{SIGHAN15}_{test}$  & 542  & 1100 & 30.6 & 7 \\
\Xhline{3\arrayrulewidth}
\end{tabular}}
\caption{Statistics of test sets, including Err\_sen (number of error sentences), Total\_sen (the total number of sentences), Avg\_len (the average length of sentences), and Max\_err (the maximum number of errors in a sentence).}
\label{test_set_data}
\vspace{-5mm}
\end{table}

We collect sentences from financial, medical, and legal domains. For the financial domain dataset, we use Public Financial Datasets\footnote{https://github.com/smoothnlp/FinancialDatasets} which are collected from online financial information. For the medical domain dataset, we use the CMeEE~\cite{2021Building} corpus from the CBLUE~\cite{zhang-etal-2022-cblue} benchmark. CMeEE is designed for NER and contains many professional medical terms. For the legal domain dataset, we employ the CAIL2018~\cite{https://doi.org/10.48550/arxiv.1807.02478, https://doi.org/10.48550/arxiv.1810.05851} dataset, which contains many criminal cases collected from verdict documents. To facilitate experimentation, each dataset was divided into three portions for training, validation, and testing. 
See Table~\ref{test_set_data} for the statistics of the test sets of these three datasets. The statistics of the test sets from three standard datasets SIGHAN13~\cite{wu-etal-2013-chinese}, SIGHAN14~\cite{yu-etal-2014-overview}, SIGHAN15~\cite{tseng-etal-2015-introduction} are also included.

In order to construct $\{X, Y\}$ samples for CSC, for a given sentence, we assign a probability of 0.7 to replace the original characters with erroneous ones. We use four methods to generate erroneous sentences: 
(1) Replace the original characters with ones that have similar pronunciations. (2) Replace the original characters with ones that have similar shapes.
(3) Try to find the maximum pinyin sequence starting from the given word that exists in the predefined pinyin-to-word set. The pinyin-to-word set collects the words that have the same pinyin sequence. We then replace the candidate words with the original words. (4) Use Hidden Markov Model (HMM)~\cite{DBLP:journals/pieee/Rabiner89} to replace the original character with the other candidate characters. According to~\cite{liu-etal-2010-visually}, phonological errors account for 83\% of the errors, while visual errors account for 48\%. Therefore, we set the ratio of the four methods to 3:3:2:2. In the implementations, we use the error text generation tool\footnote{https://github.com/liwenju0/error\_text\_gen} to help us generate the sentences. 

\begin{table*}[!t]
\centering
\resizebox{1.6\columnwidth}{!}{
\begin{tabular}{c|l|ccc|ccc|cccc|cccc}
\Xhline{3\arrayrulewidth}
\multirow{9}{*}{\textbf{Financial}} &
  \multirow{3}{*}{\textbf{Model}} & \multicolumn{6}{c|}{\textbf{Character-level}} & \multicolumn{8}{c}{\textbf{Sentence-level}} \\ \cline{3-16}
  & &
  \multicolumn{3}{c|}{\textbf{Detection}} &
  \multicolumn{3}{c|}{\textbf{Correction}} &
  \multicolumn{4}{c|}{\textbf{Detection}} &
  \multicolumn{4}{c}{\textbf{Correction}}\\ \cline{3-16}
&                       & Pre.  & Rec.  & F1       & Pre.  & Rec.  & F1 & Acc.   & Pre.  & Rec.  & F1    & Acc.   & Pre.  & Rec.  & F1    \\ \cline{2-16}
& \textbf{\underline{Supervised Methods}} \\
&FASPell*               & 60.5 & 41.0 & 48.9 & 89.9 & 60.4 & 72.3 & 43.4   & 31.9  & 22.9  & 26.7  & 41.4   & 28.0  & 20.0  & 23.3   \\
&Soft-Masked BERT*      & 88.9 & 58.1 & 70.3 & 73.1 & 47.3 & 57.4 & 53.0   & 37.3  & 34.3  & 35.7  & 43.7   & 22.5  & 20.7  & 21.6   \\
&SpellGCN*              & 90.6 & 44.3 & 59.5 & 73.8 & 40.3 & 52.1 & 45.8   & 28.9  & 25.1  & 26.9  & 39.9   & 18.9  & 16.4  & 17.5   \\
&ECSpell*               & 89.9   & 62.9  & 74.0  & 86.2  & 58.2   & 69.5 & 55.0   & 40.7  & 40.5  & 40.6  & 48.6   & 31.2  & 31.1  & 31.1   \\ \cline{2-16}
& \textbf{\underline{Unsupervised Methods}} \\
&uChecker*              & \textbf{93.7} & \textbf{83.5} & \textbf{88.3} & \textbf{92.8} & \textbf{79.9} & \textbf{85.9}  & \textbf{73.6}   & \textbf{64.7}  & \textbf{64.2}  & \textbf{64.4}  & \textbf{67.9}   & \textbf{56.3}  & \textbf{55.9}  & \textbf{56.1}  \\ \hline
\multirow{7}{*}{\textbf{Medical}}& \textbf{\underline{Supervised Methods}} \\
&FASPell*               & 77.4 & 30.8 & 44.0 & 84.7 & 56.5 & 67.8 & 46.1   & 39.4  & 21.0  & 27.4  & 43.9   & 33.1  & 17.6  & 23.0 \\
&Soft-Masked BERT*      & 83.4 & 53.3 & 65.1 & 65.8 & 43.0 & 52.0 & 56.0   & 42.9  & 37.6  & 40.1  & 46.0   & 25.0  & 21.9  & 23.3   \\
&SpellGCN*              & \textbf{90.7} & 40.9 & 56.4 & 67.8 & 39.7 & 50.1 & 46.4   & 27.4  & 23.8  & 25.5  & 40.8   & 17.7  & 15.4  & 16.4  \\
&ECSpell*               & 90.0   & 54.6  & 68.0  & 81.7  & 55.1   & 65.8 & 59.8   & 49.3  & 42.4  & 45.6  & 53.3   & 37.6  & 32.4  & 34.8   \\ \cline{2-16}
& \textbf{\underline{Unsupervised Methods}} \\
&uChecker*               & 89.3 & \textbf{78.2} & \textbf{83.3} & \textbf{87.5} & \textbf{73.1} & \textbf{79.7}   & \textbf{73.4}   & \textbf{65.3}  & \textbf{62.8}  & \textbf{64.0}  & \textbf{66.3}   & \textbf{53.7}  & \textbf{51.7}  & \textbf{52.7}\\ \hline
\multirow{7}{*}{\textbf{Legal}}& \textbf{\underline{Supervised Methods}} \\
&FASPell*               & 40.9 & 42.0 & 41.5 & 84.9 & 53.9 & 66.0 & 40.0   & 27.1  & 22.7  & 24.7  & 37.6   & 23.3  & 19.5  & 21.2 \\
&Soft-Masked BERT*     & 90.5 & 56.4 & 69.5 & 72.9 & 45.5 & 56.1 & 52.3   & 35.1  & 33.0  & 34.0  & 42.6   & 19.5  & 18.3  & 18.8\\
&SpellGCN*              & 91.8 & 43.1 & 58.7 & 72.7 & 37.9 & 49.8 & 52.2   & 40.0  & 29.3  & 33.8  & 45.3   & 25.3  & 18.6  & 21.5\\
&ECSpell*               & 91.6   & 60.9  & 73.2  & 87.9  & 57.3   & 69.4 & 55.6   & 39.6  & 39.1  & 39.3  & 49.8   & 30.8  & 30.4  & 30.6\\ \cline{2-16}
& \textbf{\underline{Unsupervised Methods}} \\
&uChecker*               & \textbf{94.5} & \textbf{84.2} & \textbf{89.0} & \textbf{93.6} & \textbf{80.6} & \textbf{86.6} & \textbf{74.0}   & \textbf{64.0}  & \textbf{64.8}  & \textbf{64.4}  & \textbf{68.3}   & \textbf{55.5}  & \textbf{56.1}  & \textbf{55.8} \\
\Xhline{3\arrayrulewidth}
\end{tabular}}
\caption{The performance of the five reproduced models on financial, medical, and legal domain test sets at the character-level and sentence-level. ``*'' denotes the model we reproduce.}
\label{domain_result_merged}
\end{table*}

\section{Experimental Settings}

\subsection{Evaluation Metrics}

For CSC models, the most widely used metrics are character-level and sentence-level precision, recall, and F1 scores. Character is the minimum component of a sentence, therefore character-level metrics can reflect the ability of a CSC model. At sentence-level, one sentence is considered to be correct only when all the wrong characters in it are successfully detected and corrected. We choose both the character-level and sentence-level metrics as evaluation metrics to fully assess the performance of the model. 

\subsection{Investigated Models}

We investigate four supervised models and one unsupervised baseline model: \textbf{FASPell}~\cite{hong-etal-2019-FASPell} is a supervised model consisting of a denoising autoencoder and a decoder. \textbf{Soft-Masked BERT}~\cite{Softmasked} is a supervised model based on BERT which modifies the masking strategy to improve model performance. \textbf{SpellGCN}~\cite{spellgcn} is a supervised model utilizing a graph convolutional network to learn character similarities from the confusion set into character classifiers. The classifiers are then processed by BERT. $\textbf{ECSpell}^{UD}$~\cite{Lv_2022} is a supervised model with an error consistent strategy in the pretraining stage of BERT to generate misspellings. The authors also utilize a user dictionary to improve the inference performance of the model. \textbf{uChecker}~\cite{li-2022-uchecker} is an unsupervised model that uses flexible masking operations to alleviate the low-source problem in the CSC.

\section{Results and Discussions}

\subsection{Reproduction of the Typical CSC Models}

\begin{table}[!t]
\centering
\resizebox{1\columnwidth}{!}{
\begin{tabular}{l|cccc|cccc}
\Xhline{3\arrayrulewidth}
\multirow{2}{*}{\textbf{Model}} & \multicolumn{4}{c|}{\textbf{Detection}} & \multicolumn{4}{c}{\textbf{Correction}} \\ \cline{2-9} 
                  & Acc. & Pre. & Rec. & F1   & Acc. & Pre. & Rec. & F1   \\ \hline
\textbf{\underline{Supervised Methods}}\\
FASPell          & 74.2 & 67.6 & 60.0 & 63.5 & 73.7 & 66.6 & 59.1 & 62.6 \\
FASPell*          & 65.5 & 50.4 & 52.9 & 51.7 & 63.2 & 45.9 & 48.2 & 47.0 \\
Soft-Masked BERT  & 80.9 & 73.7 & 73.2 & 73.5 & 77.4 & 66.7 & 66.2 & 66.4 \\
Soft-Masked BERT* & 83.0 & 74.7 & 76.9 & 75.8 & 81.8 & 72.4 & 74.5 & 73.4 \\
SpellGCN         & -    & 74.8 & 80.7 & 77.7 & -    & 72.1 & 77.7 & 75.9 \\
SpellGCN*         & -    & 75.7 & 76.9 & 76.3 & -    & 73.5 & 74.7 & 74.1 \\
ECSpell          & \textbf{86.3} & \textbf{81.1} & \textbf{83.0} & \textbf{81.0} & \textbf{85.6} & \textbf{77.5} & \textbf{81.7} & \textbf{79.5} \\
ECSpell*          & 85.2 & 77.2 & 81.3 & 79.2 & 84.5 & 75.7 & 79.7 & 77.6 \\ \hline
\textbf{\underline{Unsupervised Methods}}\\
uChecker          & 82.2 & 75.4 & 72.0 & 73.7 & 79.9 & 70.6 & 67.3 & 68.9 \\ 
uChecker*         & 77.3 & 72.5 & 64.7 & 68.4 & 75.5 & 68.3 & 61.0 & 64.5 \\
\Xhline{3\arrayrulewidth}
\end{tabular}}
\caption{The sentence-level evaluation results on SIGHAN15 test set of the original models and the corresponding models we reproduced (denoted using ``*'').}
\label{exp_result_sen}
\end{table}

\begin{table*}[!t]
\centering
\resizebox{1.6\columnwidth}{!}{
\begin{tabular}{c|cccccc|cccccccc}
\Xhline{3\arrayrulewidth}
\multirow{3}{*}{\textbf{Domain}} & \multicolumn{6}{c|}{\textbf{Character-level}}                                      & \multicolumn{8}{c}{\textbf{Sentence-level}}                                              \\ \cline{2-15} 
                        & \multicolumn{3}{c|}{\textbf{Detection}}          & \multicolumn{3}{c|}{\textbf{Correction}} & \multicolumn{4}{c|}{\textbf{Detection}}                 & \multicolumn{4}{c}{\textbf{Correction}} \\ \cline{2-15} 
                        & Pre. & Rec. & \multicolumn{1}{c|}{F1}   & Pre.      & Rec.     & F1       & Acc. & Pre. & Rec. & \multicolumn{1}{c|}{F1}   & Acc.   & Pre.  & Rec.  & F1    \\ \hline
\textbf{Financial}               & 9.3  & 84.6 & \multicolumn{1}{c|}{16.8} & 54.3      & 47.8     & 50.8     & 35.6 & 25.8 & 32.4 & \multicolumn{1}{c|}{28.8} & 28.8   & 18.4  & 23.0  & 20.4  \\ \hline
\textbf{Medical}                 & 9.4  & 89.0 & \multicolumn{1}{c|}{17.0} & 53.2      & 49.1     & 51.1     & 41.8 & 31.4 & 41.0 & \multicolumn{1}{c|}{35.5} & 35.0   & 23.2  & 30.4  & 26.3  \\ \hline
\textbf{Legal}                   & 8.1  & 83.1 & \multicolumn{1}{c|}{14.8} & 55.9      & 48.3     & 51.8     & 26.2 & 16.8 & 22.2 & \multicolumn{1}{c|}{19.1} & 22.5   & 12.6  & 16.7  & 14.4  \\ \hline
\textbf{General}                 & 6.6  & 77.3 & \multicolumn{1}{c|}{12.2} & 59.2      & 56.1     & 57.6     & 37.3 & 24.0 & 38.5 & \multicolumn{1}{c|}{29.6} & 34.8   & 20.9  & 33.5  & 25.7  \\
\Xhline{3\arrayrulewidth}
\end{tabular}}
\caption{The performance of ChatGPT on the financial, medical, legal and general test sets at the character-level and sentence-level. We use SIGHAN15 as the general test set.}
\label{result_chatgpt}
\vspace{-5mm}
\end{table*}

We reproduce all the models by ourselves for better investigation and experiments. Most models were successfully reproduced with results aligning with the original settings. However, in the case of Soft-Masked BERT, the training dataset was not publicly available, we conduct model training using a merged dataset of SIGHAN series \cite{tseng-etal-2015-introduction} and Wang-271k \cite{wang-etal-2018-hybrid}.

Since we reproduce all the models by ourselves for better investigation, we must guarantee that all the models are implemented correctly and can reproduce the original results. Therefore we conduct evaluation again on the SIGHAN15 test set for verification.
Table~\ref{exp_result_sen} show the performance of the five baseline models and our reproduced models on SIGHAN15 test set at the sentence-level and character-level respectively. From the results, we can observe that we have correctly reproduced most of the models.

\subsection{Cross-Domain Evaluation Performance of the Typical CSC Models}

After reproducing the five typical models, we test them on the three domain-specific test sets we constructed. Table~\ref{domain_result_merged} depicts the experiment evaluation results in the financial, medical, and legal domain at the character-level and the sentence-level. Based on the results in Table~\ref{domain_result_merged}, we can observe some interesting phenomena: (1) Evaluation performance indeed drops dramatically when transferring the original models to the new domain-specific tasks. This can be analyzed by comparing the sentence-level and character-level results in Table~\ref{domain_result_merged} (new domains), Table~\ref{exp_result_sen} (original domain). (2) Unsupervised model uChecker consistently  achieves good performance on the three domain-specific datasets, even though it has not learned any knowledge from the relevant domains.
(3) Among the four supervised models with pretrained models, ECSpell maintains it high performance in all the three domains.
(4) In the legal domain, we observe a decrease in the performance of most of the models at the sentence-level.
 
\subsection{Discussions of Results of the Typical CSC Models}
\label{sec:discuss}

One prominent observation is the performance drop of all the supervised models in all the three domains. Interestingly, Soft-Masked BERT, which does not rely on character similarity and pronunciation similarity, achieves higher accuracy and recall than FASPell and SpellGCN at the sentence-level. This phenomenon can be attributed to the fact that FASPell and SpellGCN, which incorporate character similarity modules, are trained on a general domain corpus. The two models tend to prioritize candidates that exhibit greater pronunciation or shape similarity with the original characters commonly encountered in daily life. Soft-Masked BERT is free of this second-time selection, so it has better performance. 

Different from other three supervised models, ECSpell still maintains good performance in the three domains. We attribute this situation to the model's use of the user dictionary. With the user dictionary, the model is more likely to select candidate words related to the specific domain, and therefore produces the better results.

By contrasting between supervised models and unsupervised models, we observe that unsupervised models has an amazing high performance in the three domains. This phenomenon can be attributed to the over-fitting of supervised models to the general domain during the training process. While pretrained models possess knowledge acquired from diverse domains, fine-tuning them using data from the general domain diminishes the utilization of this rich knowledge, thereby impacting their performance in specific domains.

\section{Performance of Large Language Model - ChatGPT}

ChatGPT\footnote{https://chat.openai.com/}, a chatbot developed by OpenAI, has gained immense popularity since its release in November 2022 due to its impressive language capabilities. We use customized prompts to evaluate the performance of ChatGPT in the financial, medical, legal and general domain. From Table~\ref{result_chatgpt}, we can observe that ChatGPT doesn't perform well across all the four tests. This deficiency primarily stems from its low precision at both the character and sentence levels. ChatGPT tends to erroneously replace correct words with incorrect ones, leading to significantly diminished precision scores. Conversely, models specifically trained on domain-specific tasks exhibit notably higher precision scores, emphasizing the importance of utilizing such specialized models.

\section{Conclusion}
In this paper, we first construct a dataset in financial, medical, and legal domain to test the domain adaptation ability of the Chinese Spelling Check models. We reproduce five state-of-the-art models and then test them on our constructed dataset. The experiment results show that unsupervised models have better adaptation ability than supervised models, and using pretrained models can effectively improve the context understanding of the model. Moreover, we also test the performance of the popular language model ChatGPT and find its weakness in CSC.

\section{Acknowledgement}
We sincerely thank all the reviewers for their thorough comments and advice. This research is supported by the National Natural Science Foundation of China (No.62106105), the CCF-Baidu Open Fund (No.CCF-Baidu202307), the CCF-Zhipu AI Large Model Fund (No.CCF-Zhipu202315), the Scientific Research Starting Foundation of Nanjing University of Aeronautics and Astronautics (No.YQR21022), and the High Performance Computing Platform of Nanjing University of Aeronautics and Astronautics.


\vfill\pagebreak



\small
\bibliographystyle{IEEEbib}
\bibliography{strings,refs}

\begin{thebibliography}{10}

\bibitem{afli-etal-2016-using}
Haithem Afli, Zhengwei Qiu, Andy Way, and P{\'a}raic Sheridan,
\newblock ``Using {SMT} for {OCR} error correction of historical texts,''
\newblock in {\em Proceedings of the Tenth International Conference on Language
  Resources and Evaluation ({LREC}'16)}, Portoro{\v{z}}, Slovenia, May 2016,
  pp. 962--966, European Language Resources Association (ELRA).

\bibitem{gao-etal-2010-large}
Jianfeng Gao, Xiaolong Li, Daniel Micol, Chris Quirk, and Xu~Sun,
\newblock ``A large scale ranker-based system for search query spelling
  correction,''
\newblock in {\em Proceedings of the 23rd International Conference on
  Computational Linguistics (Coling 2010)}, Beijing, China, Aug. 2010, pp.
  358--366, Coling 2010 Organizing Committee.

\bibitem{DBLP:journals/corr/abs-2208-01815}
Shuming Shi, Enbo Zhao, Duyu Tang, Yan Wang, Piji Li, Wei Bi, Haiyun Jiang,
  Guoping Huang, Leyang Cui, Xinting Huang, Cong Zhou, Yong Dai, and Dongyang
  Ma,
\newblock ``Effidit: Your {AI} writing assistant,''
\newblock {\em CoRR}, vol. abs/2208.01815, 2022.

\bibitem{devlin-etal-2019-bert}
Jacob Devlin, Ming-Wei Chang, Kenton Lee, and Kristina Toutanova,
\newblock ``{BERT}: Pre-training of deep bidirectional transformers for
  language understanding,''
\newblock in {\em Proceedings of the 2019 Conference of the North {A}merican
  Chapter of the Association for Computational Linguistics: Human Language
  Technologies, Volume 1 (Long and Short Papers)}, Minneapolis, Minnesota, June
  2019, pp. 4171--4186, Association for Computational Linguistics.

\bibitem{hong-etal-2019-FASPell}
Yuzhong Hong, Xianguo Yu, Neng He, Nan Liu, and Junhui Liu,
\newblock ``{FASP}ell: A fast, adaptable, simple, powerful {C}hinese spell
  checker based on {DAE}-decoder paradigm,''
\newblock in {\em Proceedings of the 5th Workshop on Noisy User-generated Text
  (W-NUT 2019)}, Hong Kong, China, Nov. 2019, pp. 160--169, Association for
  Computational Linguistics.

\bibitem{Softmasked}
Shaohua Zhang, Haoran Huang, Jicong Liu, and Hang Li,
\newblock ``Spelling error correction with soft-masked {BERT},''
\newblock {\em CoRR}, vol. abs/2005.07421, 2020.

\bibitem{spellgcn}
Xingyi Cheng, Weidi Xu, Kunlong Chen, Shaohua Jiang, Feng Wang, Taifeng Wang,
  Wei Chu, and Yuan Qi,
\newblock ``Spellgcn: Incorporating phonological and visual similarities into
  language models for chinese spelling check,''
\newblock {\em CoRR}, vol. abs/2004.14166, 2020.

\bibitem{Lv_2022}
Qi~Lv, Ziqiang Cao, Lei Geng, Chunhui Ai, Xu~Yan, and Guohong Fu,
\newblock ``General and domain adaptive chinese spelling check with error
  consistent pretraining,''
\newblock {\em {ACM} Transactions on Asian and Low-Resource Language
  Information Processing}, sep 2022.

\bibitem{li-2022-uchecker}
Piji Li,
\newblock ``u{C}hecker: Masked pretrained language models as unsupervised
  {C}hinese spelling checkers,''
\newblock in {\em Proceedings of the 29th International Conference on
  Computational Linguistics}, Gyeongju, Republic of Korea, Oct. 2022, pp.
  2812--2822, International Committee on Computational Linguistics.

\bibitem{2021Building}
H.~Zan, W.~Li, K.~Zhang, Y.~Ye, and Z.~Sui,
\newblock {\em Building a Pediatric Medical Corpus: Word Segmentation and Named
  Entity Annotation},
\newblock Chinese Lexical Semantics, 2021.

\bibitem{zhang-etal-2022-cblue}
Ningyu Zhang, Mosha Chen, Zhen Bi, Xiaozhuan Liang, Lei Li, Xin Shang, Kangping
  Yin, Chuanqi Tan, Jian Xu, Fei Huang, Luo Si, Yuan Ni, Guotong Xie, Zhifang
  Sui, Baobao Chang, Hui Zong, Zheng Yuan, Linfeng Li, Jun Yan, Hongying Zan,
  Kunli Zhang, Buzhou Tang, and Qingcai Chen,
\newblock ``{CBLUE}: A {C}hinese biomedical language understanding evaluation
  benchmark,''
\newblock in {\em Proceedings of the 60th Annual Meeting of the Association for
  Computational Linguistics (Volume 1: Long Papers)}, Dublin, Ireland, May
  2022, pp. 7888--7915, Association for Computational Linguistics.

\bibitem{https://doi.org/10.48550/arxiv.1807.02478}
Chaojun Xiao, Haoxi Zhong, Zhipeng Guo, Cunchao Tu, Zhiyuan Liu, Maosong Sun,
  Yansong Feng, Xianpei Han, Zhen Hu, Heng Wang, and Jianfeng Xu,
\newblock ``Cail2018: A large-scale legal dataset for judgment prediction,''
  2018.

\bibitem{https://doi.org/10.48550/arxiv.1810.05851}
Haoxi Zhong, Chaojun Xiao, Zhipeng Guo, Cunchao Tu, Zhiyuan Liu, Maosong Sun,
  Yansong Feng, Xianpei Han, Zhen Hu, Heng Wang, and Jianfeng Xu,
\newblock ``Overview of cail2018: Legal judgment prediction competition,''
  2018.

\bibitem{wu-etal-2013-chinese}
Shih-Hung Wu, Chao-Lin Liu, and Lung-Hao Lee,
\newblock ``{C}hinese spelling check evaluation at {SIGHAN} bake-off 2013,''
\newblock in {\em Proceedings of the Seventh {SIGHAN} Workshop on {C}hinese
  Language Processing}, Nagoya, Japan, Oct. 2013, pp. 35--42, Asian Federation
  of Natural Language Processing.

\bibitem{yu-etal-2014-overview}
Liang-Chih Yu, Lung-Hao Lee, Yuen-Hsien Tseng, and Hsin-Hsi Chen,
\newblock ``Overview of {SIGHAN} 2014 bake-off for {C}hinese spelling check,''
\newblock in {\em Proceedings of the Third {CIPS}-{SIGHAN} Joint Conference on
  {C}hinese Language Processing}, Wuhan, China, Oct. 2014, pp. 126--132,
  Association for Computational Linguistics.

\bibitem{tseng-etal-2015-introduction}
Yuen-Hsien Tseng, Lung-Hao Lee, Li-Ping Chang, and Hsin-Hsi Chen,
\newblock ``Introduction to {SIGHAN} 2015 bake-off for {C}hinese spelling
  check,''
\newblock in {\em Proceedings of the Eighth {SIGHAN} Workshop on {C}hinese
  Language Processing}, Beijing, China, July 2015, pp. 32--37, Association for
  Computational Linguistics.

\bibitem{DBLP:journals/pieee/Rabiner89}
Lawrence~R. Rabiner,
\newblock ``A tutorial on hidden markov models and selected applications in
  speech recognition,''
\newblock {\em Proc. {IEEE}}, vol. 77, no. 2, pp. 257--286, 1989.

\bibitem{liu-etal-2010-visually}
Chao-Lin Liu, Min-Hua Lai, Yi-Hsuan Chuang, and Chia-Ying Lee,
\newblock ``Visually and phonologically similar characters in incorrect
  simplified {C}hinese words,''
\newblock in {\em Coling 2010: Posters}, Beijing, China, Aug. 2010, pp.
  739--747, Coling 2010 Organizing Committee.

\bibitem{wang-etal-2018-hybrid}
Dingmin Wang, Yan Song, Jing Li, Jialong Han, and Haisong Zhang,
\newblock ``A hybrid approach to automatic corpus generation for {C}hinese
  spelling check,''
\newblock in {\em Proceedings of the 2018 Conference on Empirical Methods in
  Natural Language Processing}, Brussels, Belgium, Oct.-Nov. 2018, pp.
  2517--2527, Association for Computational Linguistics.

\end{thebibliography}

\end{document}